%% file: main_bibm24_final.tex
\documentclass[conference]{IEEEtran}
\IEEEoverridecommandlockouts
\usepackage{cite}
\usepackage{amsmath,amssymb,amsfonts}
\usepackage{algorithmic}
\usepackage{graphicx}
\usepackage{textcomp}
\usepackage{xcolor}
\usepackage{hyperref}
\usepackage{graphicx}
\usepackage{amsthm}
\usepackage{booktabs}
\usepackage{multirow}
\usepackage{makecell}
\usepackage{algorithm}
\usepackage[switch]{lineno}
\usepackage{subfigure}
\usepackage{pifont}
\usepackage{soul}

\usepackage{color, colortbl}
\newcommand{\red}[1]{\textcolor{black}{{}#1}}

\newcommand{\rred}[1]{\textcolor{black}{{}#1}}

\usepackage{mathtools}

\def\BibTeX{{\rm B\kern-.05em{\sc i\kern-.025em b}\kern-.08em
    T\kern-.1667em\lower.7ex\hbox{E}\kern-.125emX}}
\begin{document}

\title{MVKTrans: Multi-View Knowledge Transfer for Robust Multiomics Classification}

\author{\IEEEauthorblockN{Shan Cong$^{1,2}$,
Zhiling Sang$^2$,
Hongwei Liu$^1$,
Haoran Luo$^{1,2}$,
Xin Wang$^3$, \\
Hong Liang$^1$,
Jie Hao$^4$, and
Xiaohui Yao$^{1,2}$\IEEEauthorrefmark{1}}
\IEEEauthorblockA{$^1$\textit{College of Intelligent Systems Science and Engineering, Harbin Engineering University, Harbin, China}}
\IEEEauthorblockA{$^2$\textit{Qingdao Innovation and Development Center,
Harbin Engineering University, Qingdao, China}}
\IEEEauthorblockA{$^3$\textit{Department of Surgery, Chinese University of Hong Kong, Hong Kong, China}}
\IEEEauthorblockA{$^4$\textit{Institute of Medical Information, Chinese Academy of Medical Sciences \& Peking Union Medical College, Beijing, China}}
\thanks{Corresponding author: Xiaohui Yao (email: xiaohui.yao@hrbeu.edu.cn).}}

\maketitle

\begin{abstract}
The distinct characteristics of multiomics data, including complex interactions within and across biological layers and disease heterogeneity (e.g., heterogeneity in etiology and clinical symptoms), drive us to develop novel designs to address unique challenges in multiomics prediction. In this paper, we propose the multi-view knowledge transfer learning (MVKTrans) framework, which transfers intra- and inter-omics knowledge in an adaptive manner by reviewing data heterogeneity and suppressing bias transfer, thereby enhancing classification performance. Specifically, we design a graph contrastive module that is trained on unlabeled data to effectively learn and transfer the underlying intra-omics patterns to the supervised task. This unsupervised pretraining promotes learning general and unbiased representations for each modality, regardless of the downstream tasks. In light of the varying discriminative capacities of modalities across different diseases and/or samples, we introduce an adaptive and bi-directional cross-omics distillation module. This module automatically identifies richer modalities and facilitates dynamic knowledge transfer from more informative to less informative omics, thereby enabling a more robust and generalized integration. Extensive experiments on four real biomedical datasets demonstrate the superior performance and robustness of MVKTrans compared to the state-of-the-art. Code and data are available at \href{https://github.com/Yaolab-fantastic/MVKTrans}{https://github.com/Yaolab-fantastic/MVKTrans}.
\end{abstract}


\section{Introduction}

With recent technological advances in acquiring high-throughput omics data, multiomics integration is evolving as a rapidly growing research field~\cite{Han2022MultimodalDD,Leng2022GB}. 
Compared to single omics, which can only reflect a part of biological complexity from a certain perspective, integrating multiple omics types holds the capacity to capture complementary information from diverse biological layers. Multiomics integration has exhibited promising performance in various biomedical tasks, such as clinical prediction and disease subtyping~\cite{kang2022roadmap}.
Traditional approaches often involve statistical and machine learning models, which may have limited capacity to capture the complex, non-linear relationships present in multiomics data. In recent years, applying deep learning (DL) models to multiomics studies has emerged for addressing these limitations~\cite{reel2021using,kang2022roadmap,chen2020TMIpathomic,pnas2023}. 
Well-designed DL methods are capable of identifying complex patterns inherent in individual omics and integrating information from various sources, thereby enriching the analytical insight into underlying biological processes. 

Research on multiomics integration faces two main challenges: 1) the effective learning of feature representations and 2) the effective fusion of information derived from multiple omics. 
For the first challenge, different network architectures are employed to reduce redundancies and noises within high-dimensional omics data and produce meaningful and informative feature embeddings. A commonly used network is the encoder that compresses each modality into a non-linear latent space~\cite{Han2022MultimodalDD,chen2021iccv_coatt,Tang2023NC}. Autoencoder and its variants are also extensively used for learning efficient encodings of each omics type~\cite{yao2024mocat,Gao2022GLUE,Lin2022NC,Benkirane2023customics}. 
Additionally, recent methodologies often represent omics data as graphs, leveraging graph neural networks to capture complex interactions effectively \cite{2021MOGONET,Liang2024gremi,xing2022gat}. 
It is worth noting that existing approaches often integrate feature representation with downstream tasks, such as clinical classification and survival prediction.
\textit{While these task-guided strategies are designed to extract outcome-related features, their effectiveness would be influenced by factors such as disease complexity, inherent label bias, and sample heterogeneity. 
Furthermore, the intrinsic noisiness and inconsistent distribution characteristic of omics data impose limitations on the robustness and generalizability of existing methods.}

On the other hand, effective modeling of complementary and interactive information among different omics can significantly improve the discriminative potential of fusion features~\cite{Vandereyken2023NatReviewGene}. However, the exploration in this specific research domain remains relatively limited.
Regardless of the fusion stage---early, intermediate, or late---current approaches mainly use concatenation for fusing information from multiple sources~\cite{kang2022roadmap,Benkirane2023customics}, overlooking the crucial inter-omics interactions. 
Recent strategies are developed to harness omics correlation in either intermediate or late fusion stages, demonstrating superior efficacy compared to simple concatenation~\cite{Golovanevsky2022JAMIA_AD,2021MOGONET,MODILM}. 
For example, Mogonet~\cite{2021MOGONET} constructed the cross-omics tensor based on omics-specific classification probabilities and then employed the view correlation discovery network to incorporate label space omics correlations. Cross-modal attentions are also employed to capture inter-modality interactions~\cite{Golovanevsky2022JAMIA_AD,chen2021iccv_coatt}.
Alternatively, trustworthy fusion is proposed to evaluate the confidence level of each omics type and accordingly adapt to their quality differences~\cite{Han2022MultimodalDD, Cong2024TMM}.
\textit{Despite the achievements of existing efforts, the intrinsic imbalances in informativeness and complex interactions among different omics can introduce irrelevant information during the fusion process, thus, in turn, hindering robust multiomics integration.}

\begin{figure*}[!ht]
    \centering
    \includegraphics[width=0.95\textwidth]{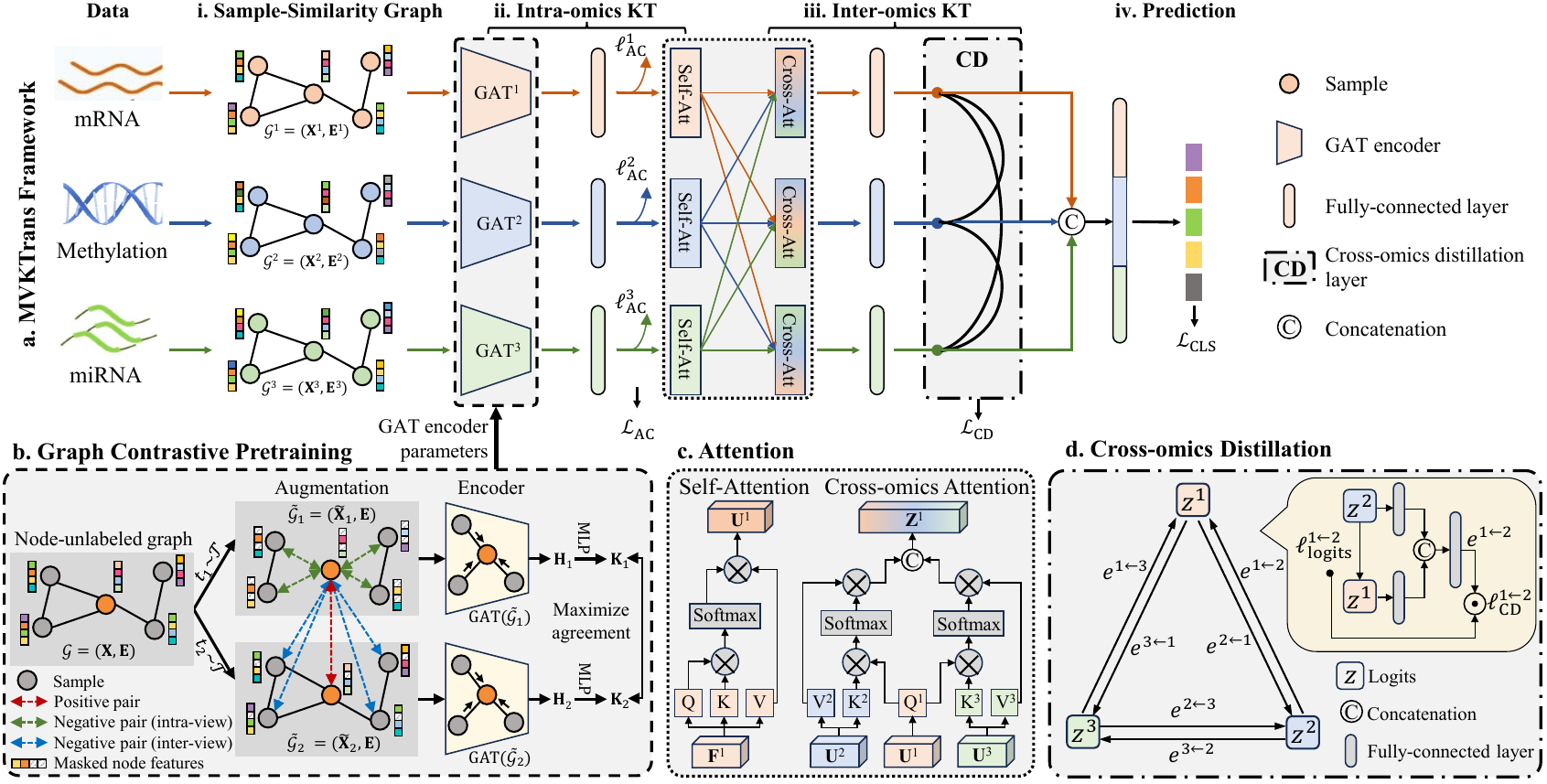}
    \caption{Framework of MVKTrans. (a) The proposed method mainly comprises the following modules. i) Sample-similarity graphs construction. ii) Intra-omics KT: Using pre-trained parameters as initialization, GATs are employed to generate representations and produce omics-specific label distributions, followed by self-attention blocks to prioritize within-omics information. Auxiliary classifiers (AC) are trained to assist in learning more representative features. iii) Inter-omics KT: Cross-omics attention is incorporated with cross-omics distillation (CD) to capture interactions and reconcile distribution disparities among omics. iv) Optimized features are concatenated to make a final prediction. (b-d) illustrates the details of graph contrastive pretraining, self-attention and cross-omics attention, and cross-omics distillation, respectively.}
    \label{fig:framework}
\end{figure*}

Based on the above observations, we propose a multi-view knowledge transfer method (MVKTrans) for robust multiomics classification as illustrated in Fig.~\ref{fig:framework}, which incorporates omics-specific pretraining (intra-view knowledge transfer (KT)) and cross-omics adaptive distillation (inter-view KT) to promote model stability and generalizability.
Specifically, during the pretraining phase, we introduce omics-specific graph contrastive learning (GCL) to initialize the model parameters trained from unlabeled data, which facilitates a foundational understanding of latent structural patterns and interrelations within each omics type.
Based on the pre-trained initializations, omics-specific graph attention networks (GAT) are built to generate initial predictions.  
Afterward, we introduce a cross-omics distillation (CD) module to facilitate adaptive knowledge transfer among disparate omics types, with a cross-omics attention module incorporated to model modality correlation and bridge the gap in label distributions. This endows the model with robustness for integrating heterogeneous omics sources.
Our main contribution can be summarized as follows:
\begin{itemize}
  \item We model the multiomics data in a multi-view knowledge transfer framework to learn and pass the intra- and inter-omics knowledge in an adaptive manner. To the best of our knowledge, this is the first work to integrate both intra- and inter-modal transfer learning for robust multimodal classification.
  \item To tackle real-world issues in multiomics data, such as label bias and omics informativeness imbalances, we specifically design an unsupervised pretraining module that employs graph contrastive learning to promote the learning of more general and unbiased representations, along with a CD module that adaptively transfers generalizable and reliable knowledge from information-rich to less informative modalities.  
  \item We conduct extensive experiments across four multiomics classification tasks to show the superiority of MVKTrans over the SOTA. Ablation and perturbation studies confirm the effectiveness and robustness of MVKTrans.
\end{itemize}

\section{The Proposed Method}\label{sec:methods}
In the following sections, we denote matrices as boldface uppercase letters and vectors as boldface lowercase ones. 
Let $m\!\!\in\!\!\![1,\dots, M]$ denotes the $m$-th omics, $\mathbf{X}^{m}\!\!=\!\!\{\mathbf{x}^m_1, \dots, \mathbf{x}^m_n\}\!\!\in\!\! \mathbb{R}^{n\times d^m}$ represents the $m$-th feature matrix, and $\mathbf{y}=[y_1,\dots,y_n]\in \mathbb{R}^{n}$ denotes the label vector, where $n$ is the number of samples and $d^m$ is the number of features.

We now detail the proposed MVKTrans. The overall architecture and important modules are illustrated in Fig.~\ref{fig:framework}.

\subsection{Omics-specific graph construction}
\label{sec:sam_sim_graph}

To formulate the functional relationships within samples and enable graph-based modeling, each omics data is transformed into a graph $\mathcal{G}^m\!\!=\!\!(\mathbf{X}^m,\mathbf{E}^m)$, where $\mathbf{X}^m\!\!\in\!\!\mathbb{R}^{n\times d^m}$ is the feature matrix, and $\mathbf{E}^m\in\mathbb{R}^{n\times n}$ is the edge matrix. 
Specifically, $\text{E}^m_{ij}$ is obtained by thresholding the adjacency matrix, which calculates sample-sample similarity:
\begin{equation}
\small
\begin{aligned}
\label{eq:adj}
    \text{E}^m_{ij}=\left\{\begin{array}{cl}
     1 & \text { if } s(\mathbf{x}^m_i, \mathbf{x}^m_j) \geq \delta \\
     0 & \text { otherwise}
    \end{array}\right.,
\end{aligned}
\end{equation}
where $s(\mathbf{x}^m_i, \mathbf{x}^m_j)=\frac{\mathbf{x}^m_i\cdot \mathbf{x}^m_j}{\left\Vert \mathbf{x}^m_i \right\Vert_2 \cdot \left\Vert \mathbf{x}^m_j \right\Vert_2 }$ computes the cosine similarity between samples $\mathbf{x}^m_i$ and $\mathbf{x}^m_j$, and $\delta$ is a hyperparameter for thresholding. We set $\delta=0.05$ across experiments. 

Consequently, $M$ sample similarity graphs $\mathcal{G}=\{\mathcal{G}^m\}_{m=1}^{M}$ are constructed to serve as model inputs, with nodes denoting samples and edges denoting the relationship between them.

\subsection{Intra-omics KT: graph contrastive pretraining}\label{sec:pretrain}
In the context of graph-structured data, GCL is widely studied, which focuses on maximizing agreement between various graph augmentations to facilitate the capture of implicit semantic information of the original data, independent of external labels~\cite{you2020nipsGCL,zhu2020GRACE}. 
Here, we utilize GCL-based pretraining to decipher the underlying molecular mechanisms inherent in each omics type. 
A typical GCL comprises four major components: graph augmentation, a backbone encoder, a projection head, and a contrastive loss function.

\textbf{Graph augmentation.} Given a graph $\mathcal{G}\!=\!(\mathbf{X},\mathbf{E})$, two corrupted graph views denoted as $\widetilde{\mathcal{G}}_1\!=\!(\widetilde{\mathbf{X}}_1,\mathbf{E})$ and $\widetilde{\mathcal{G}}_2\!=\!(\widetilde{\mathbf{X}}_2,\mathbf{E})$ are generated by randomly masking node features from $\mathcal{G}$:
\begin{equation}
\small
\begin{aligned}
\label{eq:mf}
\widetilde{\mathbf{X}}_1 = t_1(\mathbf{X},\widetilde{\mathbf{p}}_1)= 
\{ \mathbf{x}_1\odot \widetilde{\mathbf{p}}_1, \mathbf{x}_2\odot \widetilde{\mathbf{p}}_1,\dots ,\mathbf{x}_n\odot \widetilde{\mathbf{p}}_1\},\\
\widetilde{\mathbf{X}}_2 = t_2(\mathbf{X},\widetilde{\mathbf{p}}_2)=
\{ \mathbf{x}_1\odot \widetilde{\mathbf{p}}_2, \mathbf{x}_2\odot \widetilde{\mathbf{p}}_2,\dots ,\mathbf{x}_n\odot \widetilde{\mathbf{p}}_2\},
\end{aligned}
\end{equation}
where $\widetilde{\mathbf{p}}_1\!\in \!\{0,1\}^d$ and $\widetilde{\mathbf{p}}_2\!\in \!\{0,1\}^d$ are masking vectors, with each element independently sampled from Bernoulli distribution $\mathcal{B}(p_1)$ and $\mathcal{B}(p_2)$, respectively. The hyperparameters $p_1$ and $p_2$ are used to control the degree of corruption to generate $\widetilde{\mathcal{G}}_1$ and $\widetilde{\mathcal{G}}_2$. We tune $p_1$ and $p_2$ using a grid search within the range of $0.2$ to $0.8$ and set $p_1=0.3$ and $p_2=0.2$ in the implementation.

\textbf{Encoder.} 
We use multi-head GAT as the encoder to represent node features, utilizing the graph view as input to produce node embeddings, denoted as: $\mathbf{H}_{1}=\text{GAT}(\widetilde{\mathcal{G}}_1)$ and $\mathbf{H}_{2}=\text{GAT}(\widetilde{\mathcal{G}}_2)$.

\textbf{Projection head.} A multilayer perceptron (MLP) is used to map the embeddings to latent spaces $\mathbf{K}_1$ and $\mathbf{K}_2$ for applying the contrastive loss: $\mathbf{K}_1=\text{MLP}(\mathbf{H}_{1})$ and $\mathbf{K}_2=\text{MLP}(\mathbf{H}_{2})$. 

\textbf{Contrastive loss.} The contrastive objective is to make the embeddings of each node from the two views close to each other (i.e., positive pairs) while being distinguishable from the representations of other nodes (i.e., negative pairs)~\cite{zhu2020GRACE}. As shown in Fig.~\ref{fig:framework}(b), for any node $i$, its embeddings from two graph views form the positive pair $(\boldsymbol{\mu}_{i}, \boldsymbol{\upsilon}_i)$, while all other $2(n-1)$ nodes from these two views are defined as negative samples. Let $s(\boldsymbol{\mu}_{i}, \boldsymbol{\upsilon}_i)$ calculate the cosine similarity between two node embeddings. Then, for each positive pair, the contrastive loss is defined as:
\begin{equation}
\small
\!\!\ell(\mathbf{\boldsymbol{\mu}}_i, \mathbf{\boldsymbol{\upsilon}}_i)\!\!=
\!\!-\!\log\!\! 
\frac{\exp(s(\mathbf{\boldsymbol{\mu}}_i, \mathbf{\boldsymbol{\upsilon}}_i)/\tau)}
{\exp(s(\mathbf{\boldsymbol{\mu}}_i, \mathbf{\boldsymbol{\upsilon}}_i)/\tau) + 
\sum\limits_{\mathclap{\substack{j=1,j\neq i\\\mathbf{\boldsymbol{\psi}}_j\in \{\mathbf{\boldsymbol{\mu}}_j, \mathbf{\boldsymbol{\upsilon}}_j\}}}}^n {\exp(s(\mathbf{\boldsymbol{\mu}}_i, \mathbf{\boldsymbol{\psi}}_j)/\tau)}},
\end{equation}
where $\tau$ denotes a temperature parameter and is set to be $0.5$. The final loss is computed across all positive pairs:
\begin{equation}
\small
\mathcal{L}_{\text{cst}}\!\!=\!\! 
\sum\limits_{i=1}^{n} {\left( \ell(\mathbf{\boldsymbol{\mu}}_i, \mathbf{\boldsymbol{\upsilon}}_i) + \ell(\mathbf{\boldsymbol{\upsilon}}_i, \mathbf{\boldsymbol{\mu}}_i) \right)}.
\end{equation}

The parameters of GAT and MLP are updated by minimizing the contrastive loss during each training epoch. The optimized encoder parameters, denoted as $\mathbf{W}^m_{\text{GCL}}$, are transferred as initial configurations for the subsequent classification task. Using these pre-trained parameters, omics-specific GAT is fine-tuned to generate initial prediction probabilities.

\subsection{Inter-omics KT: cross-omics distillation}\label{sec:crossmodal} 
\red{To address the omics disparity issue, we introduce graph-based knowledge distillation (KD)~\cite{Luo2018GraphDF,Li2023Decoupled} to conduct adaptive cross-omics distillation. 
This integration is employed to aid the cross-attention mechanism in automatically adapting and dynamically regulating disparities during the omics fusion process. 
Traditionally, KD~\cite{hinton2015distilling} for two modalities involves transferring knowledge from the stronger to the weaker one. Considering two modalities $m_1$ and $m_2$ with the distillation direction from $m_1$ to $m_2$ (i.e., $m_1>m_2$), the training loss of KD is composed of classification loss and distillation loss: 
$\ell=\gamma_1 \ell_\text{cls} + \gamma_2\ell_\text{dsl}.$
Here, the distillation loss represents the information flowing from modality $m_1$ to $m_2$, typically quantified by the logit and/or representation distances between the source and target. }

Yet, predetermining the direction and strength of distillation is challenging when handling multiple modalities. Moreover, a diverse KD is necessary due to the heterogeneity of diseases and omics types. As shown in Fig.~\ref{fig:framework}(d), cross-omics distillation formulates a weighted, directed graph, denoted as $G=(V,\mathbf{E})$. 
The set $V$ comprises $|V|=M$ vertices with each vertex representing one omics type, while $\mathbf{E}$ is the edge matrix with each element $e^{k\leftarrow j} \in [0,1]$ denoting the distillation strength from omics $j$ to $k$. Let $\ell^{k\leftarrow j}$ denote the distillation loss from vertex $v_j$ to $v_k$ ($\ell^{k\leftarrow j} \neq \ell^{j\leftarrow k}$). Here, we employ the logits loss because we use the late-fusion strategy in the model (\red{as discussed in Sec.~\ref{sec:netarc}).} $\text{L}_1$ distance is used to measure the logits loss between source and target omics. 

Let $\mathbf{z}_i^k$ denote the logits of sample $i$ for modality $k$, the total CD loss flowing on graph $G$ can be expressed as:
\begin{small}
\begin{equation}
\label{eq:loss_gd}
\mathcal{L}_{\text{CD}}=\sum\limits_{i=1}^{n}\sum\limits_{k=1}^{M}\sum\limits_{j \in \mathcal{N}(k)} {e_i^{k\leftarrow j} \cdot \ell_i^{k\leftarrow j}},
\end{equation}
\begin{equation}
\label{eq:gd_e}
e_i^{k\leftarrow j}=\mathbf{W}_2[\mathbf{W}_1^j \mathbf{z}_i^j, \mathbf{W}_1^k \mathbf{z}_i^k],
\end{equation}
\begin{equation}
\label{eq:gd_logitloss}
\ell_i^{k\leftarrow j}=\ell^{k\leftarrow j}_{\text{logits}}=\left\Vert\mathbf{z}_i^j - \mathbf{z}_i^k\right\Vert_1,
\end{equation}
\end{small}%
where $\mathbf{W}_1$ and $\mathbf{W}_2$ are parameters to learn, $[\cdot,\cdot]$ is concatenation, $\mathcal{N}(k)$ indicates the set of source vertices for node $k$. 

\subsection{MVKTrans model architecture}\label{sec:netarc}
Now, we delineate the overall architecture of the proposed MVKTrans. As shown in Fig.~\ref{fig:framework}(a), the framework integrates multiple omics types and predicts the diagnostic status of each participant. Generally, we employ the GAT as omics encoders and use the late-fusion strategy to integrate multiomics data. More specifically, the MVKTrans is designed as follows:

\textbf{\red{GAT encoders.}} Using pre-trained parameters (denoted as $\textbf{W}_{\text{GCL}}^m$, $m\in [1,\cdots, M]$) for initialization, encoders based on the GAT architecture are fine-tuned to generate omics-specific representations \rred{$\textbf{F}^{m}=\text{GAT}(\mathcal{G}^{m}, \textbf{W}_{\text{GCL}}^m$}). 

\red{\textbf{Auxiliary classifiers.} Each auxiliary classifier (AC) uses GAT-encoded representations and sample labels to perform mono-omics prediction:} 
\begin{small}
\begin{equation}
\label{eq:gat_cls}
\widehat{\mathbf{y}^m}=\text{MLP}\left(\textbf{F}^{m}\right),
\end{equation}
\begin{equation}
\label{eq:loss_ac}    \mathcal{L}_{\text{AC}}=\sum\limits_{m=1}^{M}\ell_{\text{AC}}^{m}=\sum\limits_{m=1}^{M}
    \sum\limits_{i=1}^{n}\mathcal{L}_{\text{CE}}(y_i^m, \widehat{y_i^{m}}).
\end{equation}
\end{small}

\textbf{Omics-specific self-attention module.} 
We incorporate the self-attention mechanism (Fig.~\ref{fig:framework}(a,c)) to dynamically weigh the significance of logits from various omics types, ensuring the prioritization of the most salient omics information:
\begin{equation}
\small
  \mathbf{U}^m= 
  \text{Attention}\left(\mathbf{F}^m\mathbf{W}_Q^m, \mathbf{F}^m\mathbf{W}_K^m, \mathbf{F}^m\mathbf{W}_V^m\right).
\end{equation}

\textbf{Cross-omics attention module.} We introduce cross-omics attention to reconcile the distribution disparities between high-level heterogeneous features, thereby reinforcing further distillation. For example, omics $\mathbf{U}^m$ can be enhanced by the other omics types:
\begin{equation}
\small
  \mathbf{Z}^{m}= \big\Vert_{j=1,j\neq m}^M 
  \text{Attention}\left(\mathbf{U}^m\mathbf{W}_Q^m, \mathbf{U}^j\mathbf{W}_K^j, \mathbf{U}^j\mathbf{W}_V^j\right).
\end{equation}%

\textbf{Cross-omics distillation module.}
Details of the CD module are discussed in Sec.~\ref{sec:crossmodal}. As illustrated in \red{Fig.~\ref{fig:framework}(d)}, the distillation module is composed of FC layers and is optimized at the training time in an end-to-end manner. 

\textbf{Objective optimization.} The overall loss is composed of the auxiliary classification loss (Eq.~\ref{eq:loss_ac}), the CD loss (Eq.~\ref{eq:loss_gd}), and the cross-entropy loss of the final classification:
\begin{equation}\label{eq:loss_total}
\small
    \mathcal{L} = \lambda_1\mathcal{L}_{\text{AC}}+\lambda_2\mathcal{L}_{\text{CD}}+\mathcal{L}_{\text{Final}},
\end{equation}%
where $\lambda_1$ and $\lambda_2$ are hyperparameters for adjusting different losses. \red{We set $\lambda_1\!=\!1$ and $\lambda_2\!=\!0.005$  across all experiments.}

\section{Experiments}
\subsection{Experimental setup}
\textbf{Multiomics datasets.} Our experiments are conducted on four real-world datasets~\cite{2021MOGONET}, including ROSMAP for AD, LGG for low-grade glioma grading, BRCA for five breast cancer subtyping, and KIPAN for three renal cancer subtyping. Three omics types (mRNA, methylation, and miRNA) are available for each dataset. Details are listed in Table~\ref{tab:dataset}.
\input{tables/tab_dataset}
\input{tables/tab_sota}

\textbf{Benchmark methods.}
We compare the proposed method with 8 competitors, including three single-omics classifiers with early fusion (SVM, XGBoost, and NN), four concatenation-based methods (GRidge~\cite{van2016better}, BSPLSDA~\cite{singh2019diablo}, CF~\cite{hong2020more}, and gated multimodal fusion (GMU~\cite{arevalo2017gated})), and three methods with advanced representation and fusion designs (Mogonet~\cite{2021MOGONET}, MODILM~\cite{MODILM} and Dynamics~\cite{Han2022MultimodalDD}).

\textbf{Evaluation metrics.}
To assess the performance of methods in binary classification, we employ accuracy (ACC), F1 score (F1), and the area under the receiver operating characteristic curve (AUC). For multiclass classification, metrics include ACC, the average weighted F1 (F1-w), and the macro-averaged F1 (F1-m). Experiments are conducted five times to ascertain the mean and standard deviation. T-tests are performed to evaluate the significance of the improvements.

\textbf{Implementation details.}
We use PyTorch to develop the model and use Adam as the optimizer. In GCL pre-training, we train the model for 2000 epochs with a learning rate (lr) of 1e-3. In fine-tuning, the model is trained for 5000 epochs, with the lr of GAT set at 5e-3 and the lr for inter-omics KT set at 3e-3. We implement all experiments on an RTX 3090 GPU with 24GB memory.

\subsection{Comparison with the state-of-the-art}
Table~\ref{tab:sota} presents the comparison results on four benchmark datasets. It can be observed that 1) the proposed MVKTrans outperforms other methods in terms of three metrics, and 2) MVKTrans consistently shows significant improvements (t-test 
$p<0.05$) over the suboptimal results in the ROSMAP, LGG, and BRCA datasets. The superior performance is attributed to the particularly designed intra- and inter-omics KT modules.
In subsequent evaluations, we exclude the KIPAN dataset due to its relatively straightforward classification.

\input{tables/tab_ablation_bibm}

\subsection{Ablation study}
We conduct ablation studies to evaluate the effectiveness of the intra-omics and inter-omics KT components. We accordingly remove the GCL pretraining and replace the cross-omics distillation with fully connected NN for comparison. 
Table~\ref{tab:ablation} shows the results, from which we have the following observations: 1) incorporating GCL enhances prediction performance, affirming that unsupervised pretraining, which mitigates label bias, effectively captures latent structural patterns in omics data; 2) CD module (Fig.~\ref{fig:framework}(d)) enables compensatory distillation to enhance discriminative abilities of each omics, subsequently boosting model performance.

\begin{figure}[h]
    \centering
    \includegraphics[width=0.48\textwidth]{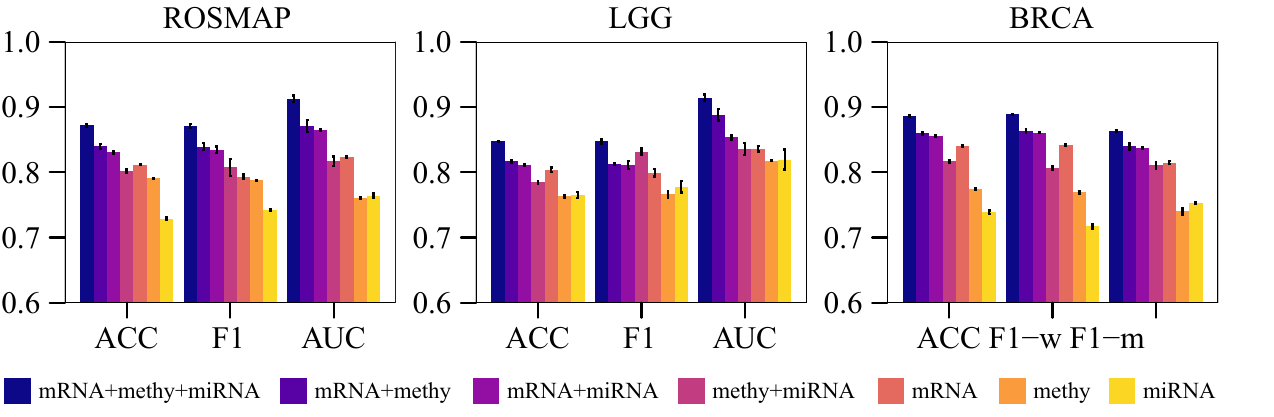}
    \caption{Performance comparison of different omics combinations.}
    \label{fig:omics123}
\end{figure}

\subsection{Evaluations of omics contributions}
Fig.~\ref{fig:omics123} shows the results of applying our model in various omics combinations.
We can observe that the integration of multiple omics surpasses their subsets, underscoring the distinct contribution of each omics and validating the capacity of our method in modeling cross-omics information.

\begin{figure}[htb]
    \centering
    \includegraphics[width=0.45\textwidth]{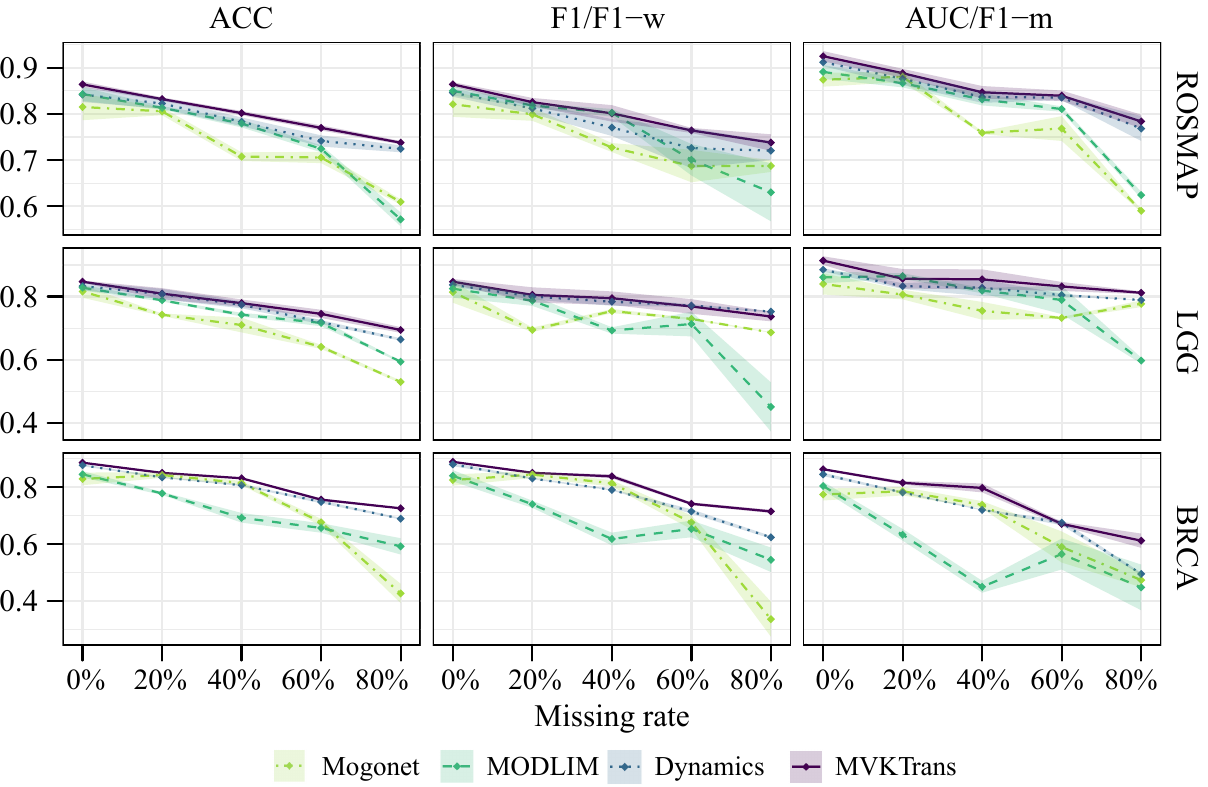}
    \caption{Robustness evaluation under different missing rates. Mean and $95\%$ confidence interval (shaded area) are shown.}
    \label{fig:robust}
\end{figure}

\subsection{Robustness study}
To evaluate robustness against data perturbation, we set varying feature missing rates (from $20\%$ to $80\%$) to the data and then compare the proposed model against others. We include the Mogonet, MODLIM, and Dynamics in comparison, given their relatively good performances. Fig.~\ref{fig:robust} illustrates that the proposed MVKTrans consistently outperforms under various missing rates, exhibiting substantial robustness at both lower and higher levels of data loss. For example, in comparison to the suboptimal Dynamics model, our method experiences a smaller decline of $5.1\%$ (vs. $9.0\%$) in F1-w on BRCA with a $40\%$ missing rate, and shows a weaker decrease of $15.3\%$ (vs. $16.9\%$) in ACC on LGG at an $80\%$ missing rate.

\input{tables/tab_biomarker}

\subsection{Biomarker identification and explanation}

Identifying biomarkers plays a crucial role in disease inference, offering essential insights that are instrumental in unraveling pathological mechanisms and facilitating novel treatments.
Feature ablation experiments~\cite{Dai_2021_WACV} are conducted to pinpoint key biomarkers, with the most significant ones presented in Table~\ref{tab:biomarker}. These findings offer promising prospects for subsequent wet-lab validation.

\section{Conclusion}
In this paper, we propose an adaptive multiomics integration framework introducing multi-view knowledge transfer for robust diagnosis classification. 
Graph contrastive pretraining is introduced to derive intra-omics patterns in an unsupervised manner. Cross-omics distillation facilitates the automatic transfer of inter-omics information from stronger to weaker modalities. Extensive experiments on benchmark datasets demonstrate the effectiveness of our approach. Concurrently, we have identified biomarkers that significantly impact disease inference, offering vital insights for elucidating pathological mechanisms and developing novel therapeutic approaches.

\section*{Acknowledgment}

This work is partly supported by the National Natural Science Foundation of China (62103116, 62102115), the Shandong Provincial Natural Science Foundation (ZR2024QF081, 2022HWYQ-093), the Fundamental Research Funds for the Central Universities (3072024GH2604), the Natural Science Foundation of Heilongjiang Province (LH2022F016).

\bibliographystyle{IEEEtran}
\bibliography{reference}

\end{document}

%% file: tables/tab_dataset.tex
\def\arraystretch{0.7}
\begin{table}[!tbh]
\footnotesize
\centering
\caption{Summary of datasets.\label{tab:dataset}}
\begin{tabular}{@{}l@{\hskip 0.2cm}|p{4cm}@{\hskip 0.2cm}|c@{\hskip 0.2cm}c@{\hskip 0.2cm}c@{}}
\toprule
\multirow{2}{*}{Dataset} & \multirow{2}{*}{Number of samples} & \multicolumn{3}{c}{Number of features} \\ \cmidrule[0.05em](lr){3-5}
 && mRNA & methy & miRNA \\
\midrule

ROSMAP & NC:169, AD:182 & 200 & 200 & 200 \\ 
\midrule

LGG & Grade2: 246, Grade3: 264 &  2,000 & 2,000 & 548 \\
\midrule

\multirow{4}{*}{BRCA}  & LuminalA: 436, LuminalB: 147, HER2-E: 46, Normal-like: 115, Basal-like: 131 &  \multirow{4}{*}{1,000}  & \multirow{4}{*}{1,000} & \multirow{4}{*}{503} \\ \midrule
KIPAN & KICH: 66, KIRC: 318, KIRP: 274 &  2,000 & 2,000 & 548 \\
\midrule
\end{tabular}
\end{table}

%% file: tables/tab_sota.tex
\begin{table*}[htb]
\footnotesize
\centering
\caption{Comparisons on four datasets. \textbf{Bold} is the best. `\textasteriskcentered{}' indicates significantly better ($p<0.05$) than suboptimal method. \label{tab:sota}}
\begin{tabular}
{p{70pt}|>{\centering\arraybackslash}p{50pt}|>{\centering\arraybackslash}p{40pt}p{40pt}p{40pt}|p{40pt}p{40pt}p{40pt}}
\toprule
\multirow{3}{*}{Method} &\multirow{3}{*}{Fusion} & \multicolumn{3}{c|}{ROSMAP} & \multicolumn{3}{c}{LGG} \\ \cmidrule(r){3-5}  \cmidrule(r){6-8}
& & ACC (\%) & F1 (\%) & AUC (\%) & ACC (\%) & F1 (\%) & AUC (\%)  \\ \midrule 
SVM & \multirow{3}{*}{early}  & 77.0±2.4 & 77.8±1.6 & 77.0±2.6 & 75.4±4.6 & 75.7±5.0 & 75.4±4.6 \\ 
XGBoost &  & 76.0±4.6 & 77.2±4.5 & 83.7±3.0 & 75.6±4.0 & 76.7±3.2 & 84.0±2.3 \\ 
NN &  & 75.5±2.1 & 76.4±2.1 & 82.7±2.5 & 73.7±2.3 & 74.8±2.4 & 81.0±3.7 \\ \midrule
GRridge & \multirow{4}{*}{intermediate}  & 76.0±3.4 & 76.9±2.9 & 84.1±2.3 & 74.6±3.8 & 75.6±3.6 & 82.6±4.4 \\ 
BSPLSDA &  & 75.3±3.3 & 76.4±3.5 & 83.8±2.1 & 68.5±2.7 & 66.2±3.0 & 73.0±2.6 \\ 
CF &  & 78.4±1.1 & 78.8±0.5 & 88.0±0.5 & 81.1±1.2 & 82.2±0.4 & 88.1±0.4 \\ 
GMU &  & 77.6±2.5 & 78.4±1.6 & 86.9±1.6 & 80.3±1.5 & 80.8±1.2 & 88.6±1.2 \\ \midrule
Mogonet & correlation & 81.5±2.3 & 82.1±2.2 & 87.4±1.2 & 81.6±1.6 & 81.4±1.4 & 84.0±2.7 \\ 
MODLIM & correlation & 84.3±1.2 & 85.0±0.8 & 89.1±1.2 & 82.8±0.7 & 82.5±2.3 & 86.1±1.5 \\ 
Dynamics & trustworthy & 84.2±1.3 & 84.6±0.7 & 91.2±0.7 & 83.3±1.0 & 83.7±0.4 & 88.5±0.4 \\ \midrule
MVKTrans (ours) & distillation & \textbf{86.4±0.5$^{*}$} & \textbf{86.4±0.4$^{*}$} & \textbf{92.5±0.9$^{*}$} & \textbf{84.7±0.3$^{*}$} & \textbf{84.7±0.7$^{*}$} & \textbf{91.4±1.1$^{*}$}  \\ 
\bottomrule
\toprule
\multirow{3}{*}{Method} &\multirow{3}{*}{Fusion} & \multicolumn{3}{c|}{BRCA} & \multicolumn{3}{c}{KIPAN} \\ \cmidrule(r){3-5}  \cmidrule(r){6-8}
 & & ACC (\%) & F1-w (\%) & F1-m (\%) & ACC (\%) & F1-w (\%) & F1-m (\%)  \\ \midrule
SVM & \multirow{3}{*}{early} & 72.9±1.8 & 70.2±1.5 & 64.0±1.7 & 99.5±0.3 & 99.5±0.3 & 99.4±0.4 \\ 
XGBoost &   & 78.1±0.8 & 76.4±1.0 & 70.1±1.7 & 99.3±0.8 & 99.3±0.8 & 98.9±1.4 \\ 
NN &   & 75.4±2.8 & 74.0±3.4 & 66.8±4.7 & 99.1±0.5 & 99.1±0.5 & 99.1±0.5 \\\midrule 
GRridge & \multirow{4}{*}{intermediate} & 74.5±1.6 & 72.6±1.9 & 65.6±2.5 & 99.4±0.4 & 99.4±0.4 & 99.3±0.4 \\ 
BSPLSDA &   & 63.9±0.8 & 52.2±1.6 & 35.1±2.2 & 91.9±1.2 & 91.8±1.3 & 89.5±1.4 \\ 
CF &   & 81.5±0.8 & 81.5±0.9 & 77.1±0.9 & 99.2±0.5 & 99.2±0.5 & 99.2±0.5 \\ 
GMU &   & 80.0±3.9 & 79.8±5.8 & 74.6±5.8 & 99.2±0.5 & 99.2±0.5 & 98.8±0.9 \\ \midrule
Mogonet & correlation & 82.9±1.8 & 82.5±1.6 & 77.4±1.7 & \textbf{99.9±0.2} & \textbf{99.9±0.2} & \textbf{99.9±0.2} \\ 
MODLIM & correlation & 84.5±0.9 & 84.0±1.6 & 80.4±1.2 & 99.2±0.8 & 99.2±0.8 & 99.2±0.8 \\ 
Dynamics & trustworthy & 87.7±0.3 & 88.0±0.5 & 84.5±0.5 & \textbf{99.9±0.2} & \textbf{99.9±0.2} & \textbf{99.9±0.2} \\ \midrule
MVKTrans (ours) & distillation & \textbf{88.6±0.2$^{*}$} & \textbf{88.9±0.1$^{*}$} & \textbf{86.3±0.2$^{*}$} & \textbf{99.9±0.2} & \textbf{99.9±0.2} & \textbf{99.9±0.2} \\ 
\bottomrule

\end{tabular}
\end{table*}

%% file: tables/tab_ablation_bibm.tex
\begin{table}[thb]
\footnotesize
\centering
\caption{Ablation study of the key components in MVKTrans.}\label{tab:ablation}
\begin{tabular}{lccccc}
\toprule
Dataset & GCL & CD  & ACC & F1 & AUC \\
\midrule
\multirow{5}{*}{ROSMAP} & $\surd$ & $\surd$ & \textbf{86.4±0.5} & \textbf{86.4±0.4} & \textbf{92.5±0.9} \\
 & $\times$ & $\surd$ & 84.2±0.3 & 84.0±0.2 & 88.0±1.0 \\
 & $\surd$ & $\times$ & 83.0±0.6 & 83.5±0.9 & 88.1±0.2 \\
  & $\times$ & $\times$ & 81.1±0.6 & 81.1±0.4 & 85.7±0.2\\\midrule
  
\multirow{5}{*}{LGG} & $\surd$ & $\surd$ & \textbf{84.7±0.3} & \textbf{84.7±0.3} & \textbf{91.4±1.1} \\
 & $\times$ & $\surd$& 83.1±0.3 & 83.4±0.5 & 88.8±0.9 \\
 & $\surd$ & $\times$  & 82.7±0.5 & 83.1±0.5 & 86.8±0.9 \\
 & $\times$ & $\times$ & 81.5±0.4 & 81.0±0.7 & 86.0±1.2\\
 \bottomrule
 \toprule
Dataset & GCL & CD & ACC & F1-w & F1-m \\
\midrule
\multirow{5}{*}{BRCA} & $\surd$ & $\surd$ & \textbf{88.6±0.2} & \textbf{88.9±0.1} & \textbf{86.3±0.2} \\
 & $\times$ & $\surd$ & 87.3±0.3 & 87.3±0.5 & 83.7±0.8 \\
 & $\surd$ & $\times$ & 85.1±0.6 & 86.1±0.7 & 82.5±0.6 \\
 & $\times$ & $\times$ & 83.9±0.2 & 84.3±0.2 & 80.5±0.6 \\
\bottomrule
\end{tabular}
\end{table}

%% file: tables/tab_biomarker.tex
\begin{table}[!htbp]
\footnotesize
  \centering
  \caption{Top 10 biomarker findings for each dataset.}
  \label{tab:biomarker}
  \begin{tabular}{@{}l@{\hskip 0.08cm}l@{\hskip 0.1cm}m{6.5cm}@{}}
\toprule
\textbf{Dataset} & \textbf{Omics} & \textbf{Top biomarkers}\\
\midrule 

\multirow{6}{*}{ROSMAP} &
mRNA & 
\textit{LINC01485, NPNT}\\\addlinespace[3pt]
~&methy & \textit{MGC40178, FZD9, PPAPDC3, HRASLS5, TRIP10, NGEF, TMEM59}\\\addlinespace[3pt]
~&miRNA & \textit{hsa-miR-132}\\
\midrule

\multirow{3}{*}{LGG} &
mRNA & 
\textit{ETV7, SCNN1B, LOC400759, FAM184B, ADAMDEC1, CENPA, IL7, CD70, NEK2}\\\addlinespace[3pt]
~&miRNA & \textit{hsa-mir-1912}\\
\midrule

\multirow{4}{*}{BRCA} &
mRNA & 
\textit{A2ML1, PTCHD1, ISL2, DSG1}\\\addlinespace[3pt]
~&miRNA & \textit{hsa-mir-194-1, hsa-mir-934, hsa-mir-3680, hsa-mir-664, hsa-mir-34a, hsa-mir-204}\\
\midrule

\multirow{6}{*}{KIPAN} &
mRNA & 
\textit{FDX2}\\\addlinespace[3pt]
~&methy & \textit{ATP5S}\\\addlinespace[3pt]
~&miRNA & \textit{hsa-mir-503, hsa-mir-891a, hsa-mir-1271, hsa-mir-577, hsa-mir-874, hsa-mir-335, hsa-mir-653, hsa-mir-1247}\\

\bottomrule   
  \end{tabular}
\end{table}